\documentclass[11pt]{article}

\usepackage{amsmath} 
\usepackage[english]{babel}
\usepackage{amsbsy}
\usepackage{amsfonts}
\usepackage{amssymb}
\usepackage{amsthm}
\usepackage[pdftex]{graphicx}
\usepackage{color}
\usepackage{float}
\usepackage{framed}
\usepackage[hyphens]{url}
\usepackage{mathtools}
\usepackage{color, colortbl}
\usepackage{xcolor}
\usepackage{mathrsfs}
\usepackage{geometry}
\usepackage[T1]{fontenc}
\usepackage[utf8]{inputenc}
\usepackage{authblk}
\usepackage{cases}
\usepackage[numbers]{natbib}
\usepackage{makecell}
\usepackage{algorithmic}
\usepackage{algorithm}
\usepackage{caption}
\usepackage{multirow}
\usepackage{subcaption}
\usepackage{diagbox}

\usepackage{hyperref}
\hypersetup{
	colorlinks=true,
	linkcolor=blue,
	filecolor=blue,
	citecolor = blue,      
	urlcolor=cyan,
}

\title{\textbf{On the Performance of Imputation Techniques for Missing Values on Healthcare Datasets}}
\author[,1]{L.O. Joel \thanks{Corresponding Author: ljoel@uj.ac.za; oluwaseyejoel@gmail.com}}
\author[,1]{W. Doorsamy \thanks{wdoorsamy@uj.ac.za}}
\author[,1]{B.S. Paul \thanks{bspaul@uj.ac.za}}
\affil[1]{Institute for Intelligent Systems\\
           University of Johannesburg\\
           South Africa}\date{}

\begin{document}

\maketitle


\begin{abstract}
\noindent Missing values or data is one popular characteristic of real-world datasets, especially healthcare data. This could be frustrating when using machine learning algorithms on such datasets, simply because most machine learning models perform poorly in the presence of missing values. The aim of this study is to compare the performance of seven imputation techniques, namely Mean imputation, Median Imputation, Last Observation carried Forward (LOCF) imputation, K-Nearest Neighbor (KNN) imputation, Interpolation imputation, Missforest imputation, and Multiple imputation by Chained Equations (MICE), on three healthcare datasets. Some percentage of missing values - 10\%, 15\%, 20\% and 25\% - were introduced into the dataset, and the imputation techniques were employed to impute these missing values.  The comparison of their performance was evaluated by using root mean squared error (RMSE) and mean absolute error (MAE). The results show that Missforest imputation performs the best followed by MICE imputation. Additionally, we try to determine whether it is better to perform feature selection before imputation or vice versa by using the following metrics - the recall, precision, f1-score and accuracy. Due to the fact that there are few literature on this and some debate on the subject among researchers, we hope that the results from this experiment will encourage data scientists and researchers to perform imputation first before feature selection when dealing with data containing missing values.
\end{abstract}

\textbf{Keywords:} Data, Missing Values, Techniques, Imputation, Healthcare

\section{Introduction}
Real-life datasets often contain some missing values or data, which pose a problem to data scientists and researchers working with them. The pattern of the missingness \cite{little2002} of these missing values could be random, that is, missing completely at random (MCAR) or missing at random (MAR). It could also be non-random, that is, not missing at random (NMAR). Some of the reasons for these missing values could be due to errors in the equipment, inappropriate pattern of data capturing, faulty sampling, damages in the specimen used, respondents' irresponsive disposition to certain information or incorrect measurements. Hence, the need to find an appropriate technique in handling these missing values so as to obtain optimal results from the analysis of the data given.

This study compares the performance of seven imputation techniques, which are Mean imputation, Median Imputation, Last Observation carried Forward (LOCF) imputation, K-Nearest Neighbor (KNN) imputation, Interpolation imputation, Missforest imputation, and Multiple imputation by Chained Equations (MICE), on three healthcare datasets, which are the breast cancer \cite{cancer2016}, the heart disease \cite{heart2018} and the pima indian diabetes \cite{diabetes2016} datasets.  Some percentage of missing values - 10\%, 15\%, 20\% and 25\% - were introduced into the datasets under the assumption of MCAR, and the imputation techniques were employed to impute these missing values. The comparison of their performance was done using two error evaluation metrics - root mean squared error (RMSE) and mean absolute error (MAE). While the evaluation metrics used to determine whether to perform selection before imputation or vice versa were the recall, precision, fi-score, and accuracy.

The rest of this paper is organised as follows. Section~\ref{sec2} talks about the datasets considered in the study and the percentage of the missing values introduced into the datasets. Section~\ref{sec3} gives the explanation of the missing data imputation techniques that will be examined in this study. Section~\ref{sec4} explains some details about feature selection and the context of it in this study. Section~\ref{sec6} describes the evaluation metrics - root mean squared error (RMSE), mean absolute percentage error (MAE), recall, precision, fi-score, and accuracy that will be used to evaluate the performance of the imputation methods. Section~\ref{sec7} gives the results and the discussion of the experiments. And lastly, the study ends with some concluding notes in Section~\ref{sec8}. 

\section{Datasets}
\label{sec2}
\subsection{Breast Cancer Dataset}
\begin{figure}[!htb]
	\begin{center}
		\includegraphics[width=2.5in]{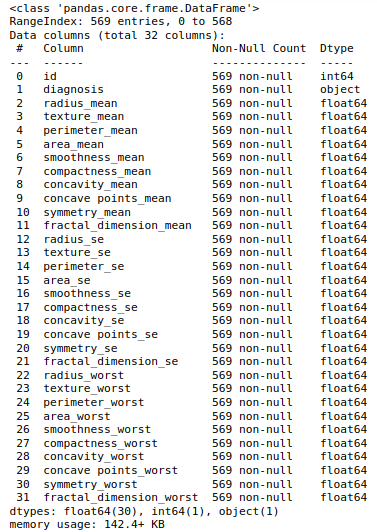}
		\caption{The Breast Cancer Dataset Information}
		\label{cancerData}
	\end{center}
\end{figure}

\begin{figure}[!htb]
	\begin{center}
		\includegraphics[width=3.5in]{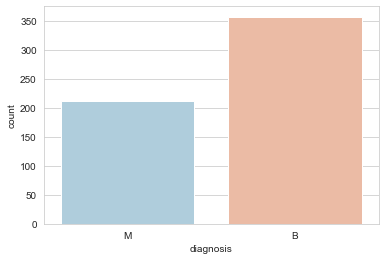}
		\caption{The Distribution of the Target Variable}
		\label{cancerData2}
	\end{center}
\end{figure}

Breast cancer is the most common and leading cause of deaths in females in many countries of the world. The first common symptom of breast cancer is a growth or  lump in the breast \cite{adam2021b}. This lump can either be cancerous (malignant) or non-cancerous (benign), a doctor has to be consulted for appropriate diagnosis. The dataset for this breast cancer is taken from kaggle database \cite{cancer2016}. Figure~\ref{cancerData} shows the different features and their data type in the dataset~\cite{cancer2016}. While Figure~\ref{cancerData2} shows the distribution of the target feature, called "diagnosis", in the dataset. "M" stands for malignant and "B" stands for benign.

\subsection{Diabetes Mellitus Dataset}
Diabetes mellitus, commonly called diabetes, is a disease that hinders the body from making enough insulin in order to move sugar from the blood into the cells that will make use of it for energy, thereby causing high blood sugar. This high blood sugar can cause damage to kidneys, eyes, nerves, and other organs in the body. Diabetes can be any of these three types: type 1, type 2, and gestational diabetes.

\begin{figure}[!ht]
	\begin{center}
		\includegraphics[width=2.5in]{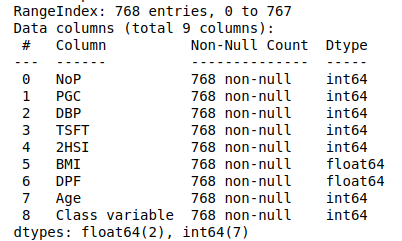}
		\caption{The Diabetes Dataset Information}
		\label{diabetesData2}
	\end{center}
\end{figure}

\begin{figure}[!ht]
	\begin{center}
		\includegraphics[width=3.5in]{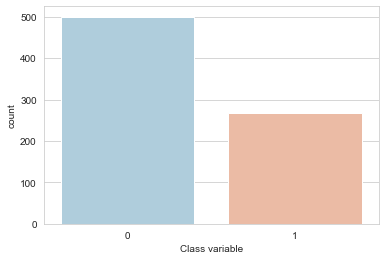}
		\caption{The Distribution of the Target Variable}
		\label{diabetesData}
	\end{center}
\end{figure}

The diabetes dataset used in this study contains no missing values, but some percentages of missing values were later introduced into the dataset so as to evaluate the performance of the various imputation techniques. The dataset is taken from the popular kaggle database \cite{diabetes2016}. It contains 768 features (rows) and 9 columns, which include the target or dependent feature (called the Class variable), see Figure~\ref{diabetesData2}. The distribution of the class variable can be seen in Figure~\ref{diabetesData}, where $1$ represents the presence of diabetes and $0$, otherwise.

\subsection{Heart Disease Dataset}
Heart disease can be referred to as any adverse condition affecting the heart. There are different types of heart disease \cite{adam2021} but the most common type is the coronary artery disease. This is when the arteries that supply blood to the heart is clogged, which in turn reduces blood supply, oxygen and nutrients needed for the proper functioning of the heart. It is the leading cause of death in the United States of America \cite{murphy2018}. This makes heart disease a major concern in healthcare and any missing values in the dataset could adversely affect the outcome of any machine learning algorithm employed in its prediction. 
\begin{figure}[!htb]
\begin{center}
\includegraphics[width=2.5in]{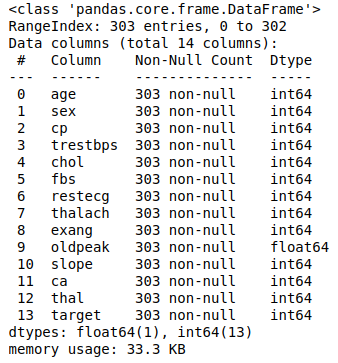}
\caption{The Heart Disease Dataset Information}
\label{heartData}
\end{center}
\end{figure}

\begin{figure}[!htb]
\begin{center}
\includegraphics[width=3.5in]{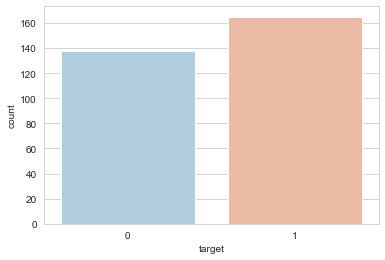}
\caption{The Distribution of the Target Variable}
\label{heartData2}
\end{center}
\end{figure}

The heart disease dataset is taken from the popular dataset database \cite{heart2018}. It contains 303 features (rows) and 14 columns, which include the target or dependent feature, see Figure \ref{heartData}. In the distribution of the target variable in Figure~\ref{heartData2}, $0$ represents presence of heart disease while $1$ stands for the  absence of it.

\section{Missing Data Imputation Techniques}
\label{sec3}
This section discusses some selected imputation techniques that will be used in this study. Additional notes on the selected techniques and/or other techniques used in handling missing values could be found in the following literature \cite{de2003, ding2012, hadeed2020, jadhav2019, kang2013}. 

\paragraph{Mean Imputation:} It is also called mean substitution. This method is very popular among researchers for missing data imputations. It replaces a missing variable in a feature with the mean of the non-missing variables in the same feature \cite{ochieng2020,sim2015}. While it is easy to understand and compute, it leads to underestimation of standard errors. 

\paragraph{Median Imputation:} Median imputation replaces all the occurrences of missing values with the middle value of the non-missing variables in the same feature \cite{hadeed2020}. It is suitable for continuous and discrete numerical variables only. Although, it is easy to implement, but it could cause distortion in the variable distribution and variance.

\paragraph{Last Observation Carried Forward Imputation:} Last Observation Carried Forward (LOCF) imputation method, which is commonly used in the analysis of clinical results when the dataset are longitudinal, fills in the missing values of an independent feature with the last non-missing observation of the feature \cite{hadeed2020, lachin2016}. Hence, this method works on the assumption that the response at the last observed value remains constant. 

\paragraph{K-Nearest Neighbor Imputation:}
The K-nearest neighbor (KNN) imputation method imputes missing values in a feature by finding the observations in the dataset closest to the observation which contains the missing values and averages these nearby points to substitute the missing values. The KNN imputation method appears to be robust and effective in missing values imputations \cite{ding2012,jadhav2019,zhang2012}. KNN's configuration often requires selecting a distance measure (such as the Hamming, Euclidean, or the  Manhattan distance) and the number of neighbors, $k$, that will be used to predict each missing value.

\paragraph{Interpolation Imputation:} This method fills the missing values with incrementing or decrementing values by performing linear, quadratic or cubic interpolation imputation on the dataset containing the missing values \cite{noor2014,noor2015,kornelsen2014}.  The author in \cite{noor2015} compared the performance of linear interpolation imputation method with mean imputation for estimating the missing values in environmental data and found the linear interpolation method to perform better than the mean method in the three evaluating metrics used.

\paragraph{Missforest Imputation:} Missforest is an imputation algorithm that uses random forest for the imputation of missing data \cite{hong2020,stekhoven2012}. Missforest imputation first fills the missing values using the mean or mode, then it fits a random forest algorithm on the observed data in order to predict the missing data. This process is performed iteratively until a stopping criterion is met or a maximum number of iterations is attained. These multiple iterations allows the random forest algorithm to improve on the quality of the trained data for the final imputation.

\paragraph{Multiple Imputation by Chained Equations:}
The Mean, LOCF, KNN, LinR, and SR methods described above only create a single value for imputing each missing value. However,  multiple imputation method creates multiple values for the imputation of a missing value in order to have different plausible imputed datasets \cite{wulff2017,azur2011,pampaka2016}. It allows for the reflection of sampling variability which is lacking in the single imputation methods. One of the commonly used multiple imputation algorithms, among others, is the multiple imputation by chained equations (MICE).

Python programming language is employed for these imputation techniques listed above with the following two python packages - \texttt{imputena} \cite{imputena2020} and \texttt{missingpy} \cite{missingpy2018}. The packages permit the automated, as well as the customized treatment of missing values in any given dataset.

\section{Feature Selection}
\label{sec4}
A given dataset often contains a plethora of features. However, in some cases or most cases, not all the features are useful in building a predictive machine learning model. Also, wrong selection of the features might make the prediction results worse. Hence, the need for right feature selection to be done in order to build an optimal machine learning model. 

Feature selection is necessary in ML so as to reduce the curse of dimensionality and to build a model that is simple and explainable. It aims to choose a subset of the features in a given dataset, known as the relevant or best features, by removing irrelevant and redundant ones \cite{Saurav2016}.  This can be done in three broad categories, namely (1) Filter Method, (2) Wrapper Method and (3) Embedded Method. See the following literature \cite{chandrashekar2014,subbiah2021,kumar2014,saeys2007}.

This study employs the sequential forward selection (SFS) algorithm to select a subset of the features in each dataset that are most relevant to the each problem. SFS is a family of greedy search algorithm that eliminates or adds features based on a given classifier performance metric.

\section{Evaluation Metrics}
\label{sec6}
This section discusses the metrics used for the evaluation of the performance of the imputation methods. The metrics are the RMSE, MAE, recall, precision, f1-score, and accuracy. A brief explanation of each is given below. 

\subsection*{Root Mean Square Error}
The root mean square error (RMSE) represents the quadratic mean of the differences between the imputed and observed data. It is shown in Equation~(\ref{rmse}) and it is one of the most commonly used metrics in the literature \cite{hadeed2020, schmitt2015, li2014}. The value of RMSE is always non-negative and a lower value is better than an higher value.  
\begin{equation}
\text{RMSE} = \sqrt{\frac{1}{n} \sum_{i=1}^{n} \left( X_{i}^{obs} - X_{i}^{imputed} \right)^2 }.
\label{rmse}
\end{equation}

\subsection*{Mean Absolute Error}
The mean absolute error (MAE) is the mean absolute difference between the actual and the imputed data. It is also one of the most commonly used metrics in the literature \cite{botchkarev2018b}. The formula for MAE is shown in Equation~(\ref{mae}). It has an advantage of the absolute value used in the formula, and a lower value is preferable to a larger value. It is robust to outliers \cite{christian2018}.
\begin{equation}
\text{MAE} =  \frac{1}{n} \sum_{i=1}^{n} |X_{i}^{obs} - X_{i}^{imputed}|
\label{mae}
\end{equation}

\subsection*{Recall}
Recall, also called sensitivity or True positive rate, is the percentage of the total relevant outcome correctly predicted or classified by the algorithm. It gives the measure of how accurately our model is able to identify those patients that have the disease (either has malignant breast cancer or diabetic or heart disease). We need to predict as many of them as possible, hence a high recall value is needed. That is, we need a low value of false negative. The formula for recall is given in Equation~(\ref{recall}).

\begin{equation}
\text{Recall} = \frac{\text{True Positive}}{\text{True Postive + False Negative}}
\label{recall}
\end{equation}

\subsection*{Precision}
Precision, which is also called Positive Predictive Value (PPV), is defined as the fraction of positive predictions that are actually correct. The formula is given in Equation~(\ref{precission}). This means when precision is improved, typically recall will be reduced and vice versa.
\begin{equation}
\text{Precission} = \frac{\text{True Positive}}{\text{True Positive + False Positive}} 
\label{precission}
\end{equation}

\subsection*{F1-Score}
F1-score is the harmonic mean of recall and precision. 
Harmonic mean is used, instead of arithmetic or geometric mean, because it equalizes the weights of the recall and precision. F1-score shows the predictive power of the classification algorithm or model. A Higher F1-score value shows a higher predictive power of the model. The formula for F1-score is shown in Equation~(\ref{f1score}).
\begin{equation}
\text{F1-score} = 2 * \frac{\text{Recall * Precision}}{\text{Recall + Precision}}
\label{f1score}
\end{equation}

\subsection*{Accuracy}
Accuracy is the fraction of the total number of predictions that were correctly predicted. In simple terms, it is the fraction of the predictions that the model got correctly. The formula for accuracy is given in Equation~(\ref{accuracy}).
\begin{equation}
\text{Accuracy} = \frac{\text{Correctly  Predicted}}{\text{Total Predictions}}
\label{accuracy}
\end{equation}

\section{Results and Discussion}
\label{sec7}
First, it is observed that the target feature shown in the diabetes dataset (Figure~\ref{diabetesData}) seems imbalanced. To avoid bias by the classification algorithm, oversampling method was used for this, which brings the total observations in the dataset to $1000$. It was $768$ before the sampling method was applied. The other two datasets - breast cancer and heart disease - do not need to be balanced. Also, for the breast cancer disease dataset, the target feature, called "diagnosis", was encoded into $1$ and $0$ using the "LabelEncoder" transformer. Hence, in Figure \ref{cancerData2}, 'M' is encoded to $1$ and 'B' is encoded to $0$. The other two datasets - diabetes and heart disease - do not require this process. The missing values were imputed using the imputation methods. For each percentage of missing values introduced into the dataset, we perform imputations using the various methods. 

\subsection{Performance of the Imputation Methods}

\subsubsection{Breast Cancer}
\begin{figure}[H]
	\begin{center}
		\includegraphics[width=0.49\textwidth]{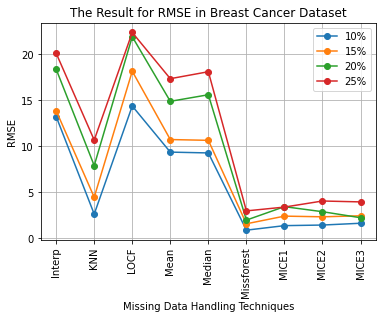}
		\hfil
		\includegraphics[width=0.49\textwidth]{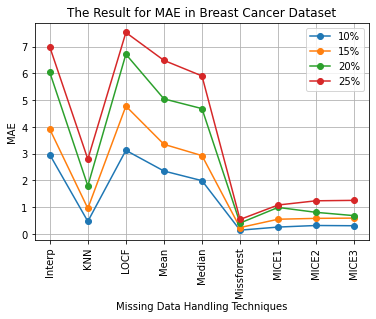}
	\end{center}
	\caption{The Errors for each of the Missing Data Handling Technique Imputations in Breast Cancer Dataset: RMSE (left side) and MAE (right side). }
	\label{cancer_error}
\end{figure}

Figure~\ref{cancer_error} gives the RMSE and MAE of the missing data handling technique imputations for 10\%, 15\%, 20\% and 25\% missing values respectively. The results given in the figure shows that Missforest algorithm has the lowest errors for both RMSE and MAE. This demonstrates that it performs best compare to other imputation methods used. Following the Missforest algorithm, is the MICE algorithm, which has lowest errors among the remaining imputation methods excluding the Missforest algorithm. The next algorithm with the lowest errors following Missforest and MICE is the KNN algorithm, for both RMSE and MAE. The LOCF method has the highest errors for both RMSE and MAE. Hence, it performs worst than the others for data missing imputation on the breast cancer data. The order of performance of the missing data imputation methods (see Figure~\ref{cancer_error}), from the lowest error to the highest error, is Missforest, MICE, KNN, Median, Mean, Interpolation, and LOCF. This order is the same for both RMSE and MAE.
The order of performance for each imputation method is the same for all percentages - 10\%, 15\%, 20\% and 25\% - of the missing values. For instance, the order of performance for 10\% missing values in the RMSE (breast cancer dataset) is Missforest, MICE, KNN, Median, Mean, Interp, and LOCF. This order is also the same for 15\%, 20\% and 25\% missing values. This scenario is played out for both the results in RMSE and MAE on breast cancer dataset.

\subsubsection{Diabetes}
\begin{figure}[H]
	\begin{center}
		\includegraphics[width=0.49\textwidth]{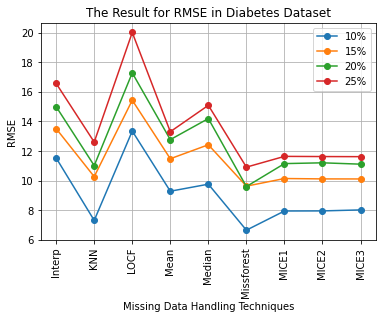}
		\hfil
		\includegraphics[width=0.49\textwidth]{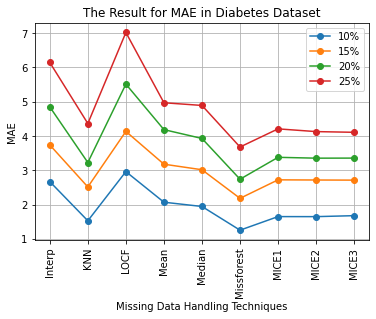}
	\end{center}
	\caption{The Errors for each of the Missing Data Handling Technique Imputations in Diabetes Dataset: RMSE (left side) and MAE (right side).}
	\label{diabetes_error}
\end{figure}

The results of the imputation errors on diabetes dataset is shown in Figure~\ref{diabetes_error}, which gives the RMSE and MAE of the imputations for 10\%, 15\%, 20\% and 25\% missing values respectively. Here, the Missforest algorithm also has the lowest errors for both RMSE and MAE. Thus, it is the best performing algorithm on the diabetes dataset. The next imputation method with the lowest errors, aside the Missforest algorithm, is the KNN method. The KNN imputation method outperforms all other remaining imputation methods including the MICE, making it the second best. The third performing imputation method is the MICE. Again, the method with the highest errors for both RMSE and MAE is the LOCF imputation method. The order of performance for the missing data imputation methods (see Figure~\ref{diabetes_error}), from the one with the lowest RMSE to the one with the highest RMSE is Missforest, KNN, MICE, Mean, Median, Interpolation, and LOCF. While the order for MAE is Missforest, KNN, MICE, Median, Mean, Interpolation and LOCF.
As described for breast cancer dataset, the order of performances of each of the imputation methods in the diabetes dataset is also maintained for all the percentages - 10\%, 15\%, 20\% and 25\% - of the missing values for the results in RMSE and MAE.

\subsubsection{Heart Disease}
\begin{figure}[H]
	\begin{center}
		\includegraphics[width=0.49\textwidth]{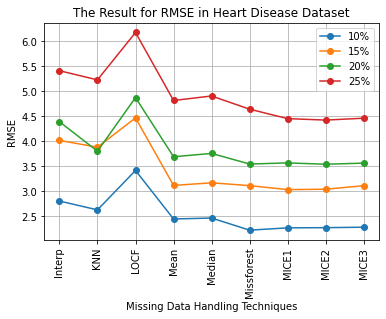}
		\hfil
		\includegraphics[width=0.49\textwidth]{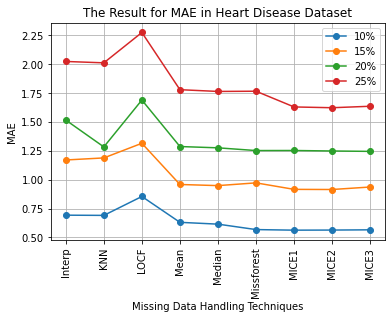}
	\end{center}
	\caption{The Errors for each of the Missing Data Handling Technique Imputations in Heart Disease Dataset: RMSE (left side) and MAE (right side).}
	\label{heart_error}
\end{figure}

The results for the imputations in heart disease dataset is shown in Figure~\ref{heart_error}. The RMSE and MAE for 10\%, 15\%, 20\% and 25\% missing values is slightly different from what was seen in the previous two datasets. Although the Missforest imputation method still maintains the lowest error for RMSE but has around the same error values with MICE for MAE. The errors in KNN imputation method is higher than both Mean and Median imputations, which is not the case in the breast cancer and diabetes datasets. The order, from the lowest error to the highest error for the RMSE is: Missforest, MICE, Median/Mean, KNN, Interpolation, and LOCF. While the order, from the lowest error to the highest error for the MAE is: Missforest/MICE, Median, Mean, KNN/Interpolation, LOCF.
The order of performances of each of the imputation methods in the heart disease dataset is only maintained for 10\%, 15\%, and 20\% of missing values in RMSE and for 10\% and 20\% of missing values in MAE. In RMSE, the order of the performances for 10\%, 15\% and 20\% of missing values is Missforest, MICE, Mean, Median, KNN, Interp, andd LOCF. However, for 25\% of missing values, the order of performance is MICE, Missforest, Mean, Median, KNN, Interp, and LOCF. Also, in MAE, the order of performances for 10\% and 20\% of missing values is Missforest, MICE, Median, Mean, KNN, Interp, and LOCF. While the order of perforamnces for 15\% of missing values is MICE, Median, Mean, Missforest, Interp, KNN and LOCF. And the order of performance for 25\% of missing values is MICE, Missforest, Median, Mean, KNN, Interp and LOCF.

\subsection{Feature Selection Before Imputation or Vice Versa}
The best two imputation methods from the previous subsection were selected to determine whether it is better to do feature selection before imputation or to do imputation before feature selection. These two methods are Missforest and MICE. Random Forest algorithm was used for the classification while the best two methods mentioned earlier were used for imputing the missing values. The experiment was done on the following percentages - 15\% and 20\% - of the missing values. And the recall, precision, f1-score and accuracy were used to evaluate the performances.
\begin{figure}[!htb]
	\begin{center}
		\includegraphics[width=0.9\textwidth]{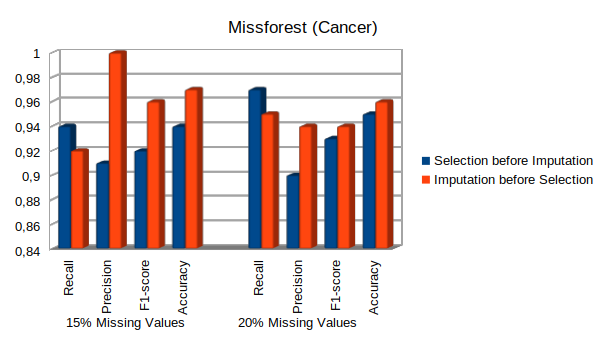}
		\hfill
		\includegraphics[width=0.9\textwidth]{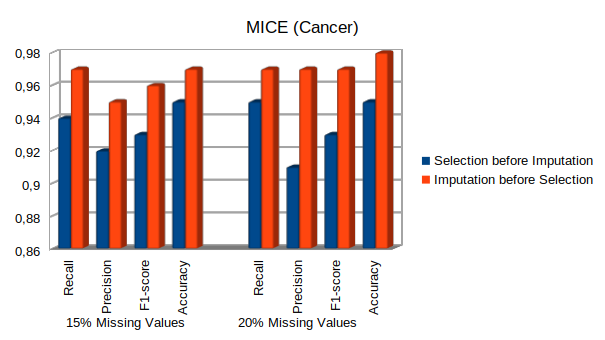}
		\caption{The performance of the Classification Algorithm for Feature Selection before Imputation Versus Imputation Before Feature Selection (Breast Cancer): Missforest (up) and MICE (down)}
		\label{cancer_compare}
	\end{center}
\end{figure}
Firstly, the results for breast cancer dataset, Figure~\ref{cancer_compare}, showed that the performance of the classification algorithm when the imputation is done before feature selection is better. However, it is observed that the recall score of the Missforest imputation method (see the figure at the top of Figure~\ref{cancer_compare}) for both 15\% and 20\%, suggests otherwise. There is a higher value for recall when feature selection is done before imputation. Hence, the performance for Missforest (Breast Cancer) classification when feature selection is performed before imputation can be rated $1/4$ for both 15\% and 20\% of missing values.

While the results of the  performance of Missforest (Breast Cancer) when imputation is done before feature selection can be rated $3/4$ for both 15\% and 20\% of missing values. On the other hand, the results for MICE (Breast Cancer) classification for feature selection before imputation can be rated $0/4$ while that of imputation before feature selection can be rated $4/4$ for both 15\% and 20\% of missing values. Hence, for the breast cancer dataset, we can conclude, from the experiment, that it is better to perform imputation before feature selection.
\begin{figure}[!htb]
	\begin{center}
		\includegraphics[width=0.9\textwidth]{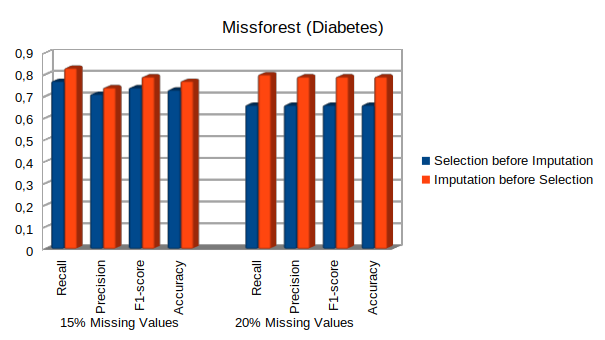}
		\hfill
		\includegraphics[width=0.9\textwidth]{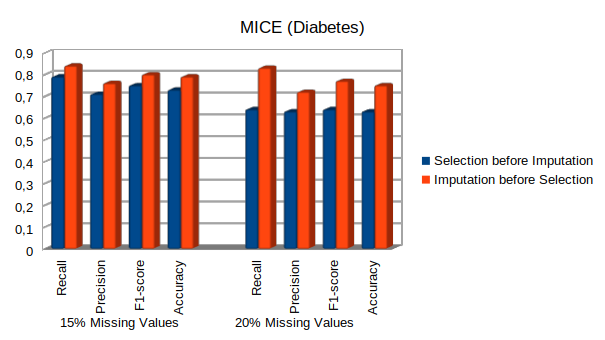}
		\caption{The performance of the Classification Algorithm for Feature Selection before Imputation Versus Imputation Before Feature Selection (Diabetes): Missforest (up) and MICE (down) }
		\label{diabetes_compare}
	\end{center}
\end{figure}

Secondly, the results for the diabetes dataset (Figure \ref{diabetes_compare}) also showed that the performance of the classification algorithm is better when the imputation of missing values is done before feature selection in a given dataset. In Missforest (Diabetes) classification, the performance when imputation is done before feature selection can be rated $4/4$ for both 15\% and 20\% missing values. While the performance when feature selection is performed before imputation can be rated $0/4$ in both missing percentages. 

Also, in MICE (Diabetes) classification results, the performance when imputation is done before feature selection can be rated $4/4$ for both 15\% and 20\% missing values. While the converse process also gives $0/4$ in both missing percentages. Hence, the results suggest that it is better to perform imputation before the feature selection step in working with any given dataset.
\begin{figure}[!htb]
	\begin{center}
		\includegraphics[width=0.9\textwidth]{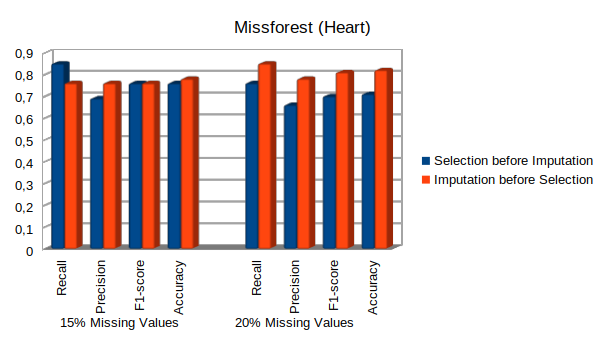}
		\hfill
		\includegraphics[width=0.9\textwidth]{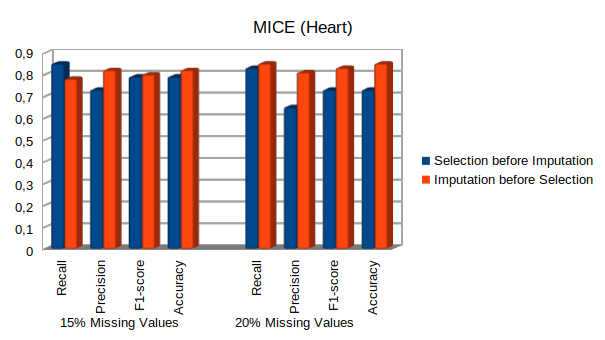}
		\caption{The performance of the Classification Algorithm for Feature Selection before Imputation Versus Imputation Before Feature Selection (Heart Disease): Missforest (up) and MICE (down) }
		\label{heart_compare}
	\end{center}
\end{figure}
Lastly, in Figure~\ref{heart_compare}, the classification performance on heart disease dataset for imputing the missing values before feature selection is, again, observed to be better than when feature selection is done before imputation. However, the recall score for 15\% missing values in both Missforest (Heart Disease) and MICE (Heart Disease) showed otherwise. However, this is just one out of the four metrics used.

In both the Missforest (Heart Disease) and MICE (Heart Disease) classification, the performance when imputation is done before feature selection can be rated $3/4$ for 15\% missing values. While the performance when feature selection is performed before imputation can be rated $1/4$ for 15\% missing values. However, the classification performance rating in both Missforest (Heart Disease) and MICE (Heart Disease) for 20\% when imputation is done before feature selection is $4/4$ while the converse procedure is $0/4$.

\clearpage

\section{Conclusion}
\label{sec8}
This study aimed to achieve two things: (1) to evaluate the performance of seven missing values imputation methods on three healthcare datasets, namely the breast cancer, diabetes mellitus and heart disease datasets. (2) to determine whether it is better to impute missing values before performing feature selection on a given dataset or to perform feature selection on the dataset before imputing the missing values. 

To achieve the first objective, the RMSE and MAE were used as evaluation metrics for the performances of the missing data handling techniques. Lower value of both RMSE and MAE demonstrates better performance of the methods. Missforest imputation method got the lowest error for both RMSE and MAE in most of the percentages of the missing values introduced into the three healthcare datasets. Hence, it performed the best among the imputation methods. Next in performance, is the MICE imputation. In a similar study carried out by Wu et al.\cite{wu2019}, MICE was one of the two suggested best imputation methods that could perform better with small scale database.

For the second objective, random forest algorithm was used for the classification predictions and the metrics used were the recall, precision, f1-score, and accuracy. The experiments were conducted using the two best imputation methods - Missforest and MICE - from the results of the performances of the seven imputation methods used in the previous experiments. The results, from the second experiments, show that it is better to impute the missing values first in a given healthcare dataset before performing feature selection than to perform feature selection before imputation. 


\bibliographystyle{unsrt}
\bibliography{paper3Ref.bib}

\begin{thebibliography}{10}

\bibitem{little2002}
Roderick J~A Little and Donald~B Rubin.
\newblock {\em Statistical Analysis with Missing Data}.
\newblock John Wiley \& Sons, Inc., USA, 2002.

\bibitem{cancer2016}
Kaggle.
\newblock Breast cancer wisconsin (diagnostic) data set.
\newblock \url{https://www.kaggle.com/uciml/breast-cancer-wisconsin-data},
  2016.

\bibitem{heart2018}
Kaggle.
\newblock Heart disease uci.
\newblock \url{https://www.kaggle.com/ronitf/heart-disease-uci}, 2018.

\bibitem{diabetes2016}
Kaggle.
\newblock Pima indians diabetes database.
\newblock \url{https://www.kaggle.com/uciml/pima-indians-diabetes-database},
  2016.

\bibitem{adam2021b}
Adam Felman.
\newblock What to know about breast cancer.
\newblock \url{https://www.medicalnewstoday.com/articles/37136#symptoms}, 2021.

\bibitem{adam2021}
Adam Felman.
\newblock Everything you need to know about heart disease.
\newblock \url{https://www.medicalnewstoday.com/articles/237191}, 2021.

\bibitem{murphy2018}
Sherry~L Murphy, Jiaquan Xu, Kenneth~D Kochanek, and Elizabeth Arias.
\newblock Mortality in the united states, 2017.
\newblock 2018.

\bibitem{de2003}
Edith~D De~Leeuw, Joop~J Hox, and Mark Huisman.
\newblock Prevention and treatment of item nonresponse.
\newblock {\em Journal of Official Statistics}, 19:153--176, 2003.

\bibitem{ding2012}
Yaohui Ding and Arun Ross.
\newblock A comparison of imputation methods for handling missing scores in
  biometric fusion.
\newblock {\em Pattern Recognition}, 45(3):919--933, 2012.

\bibitem{hadeed2020}
Steven~J Hadeed, Mary~Kay O'Rourke, Jefferey~L Burgess, Robin~B Harris, and
  Robert~A Canales.
\newblock Imputation methods for addressing missing data in short-term
  monitoring of air pollutants.
\newblock {\em Science of The Total Environment}, page 139140, 2020.

\bibitem{jadhav2019}
Anil Jadhav, Dhanya Pramod, and Krishnan Ramanathan.
\newblock Comparison of performance of data imputation methods for numeric
  dataset.
\newblock {\em Applied Artificial Intelligence}, 33(10):913--933, 2019.

\bibitem{kang2013}
Hyun Kang.
\newblock The prevention and handling of the missing data.
\newblock {\em Korean journal of anesthesiology}, 64(5):402, 2013.

\bibitem{ochieng2020}
Fredrick Ochieng’Odhiambo.
\newblock Comparative study of various methods of handling missing data.
\newblock {\em Mathematical Modelling and Applications}, 5(2):87, 2020.

\bibitem{sim2015}
Jaemun Sim, Jonathan~Sangyun Lee, and Ohbyung Kwon.
\newblock Missing values and optimal selection of an imputation method and
  classification algorithm to improve the accuracy of ubiquitous computing
  applications.
\newblock {\em Mathematical problems in engineering}, 2015, 2015.

\bibitem{lachin2016}
John~M Lachin.
\newblock Fallacies of last observation carried forward analyses.
\newblock {\em Clinical trials}, 13(2):161--168, 2016.

\bibitem{zhang2012}
Shichao Zhang.
\newblock Nearest neighbor selection for iteratively knn imputation.
\newblock {\em Journal of Systems and Software}, 85(11):2541--2552, 2012.

\bibitem{noor2014}
MN~Noor, AS~Yahaya, Nor~Azam Ramli, and Abdullah Mohd~Mustafa Al~Bakri.
\newblock {\em Filling missing data using interpolation methods: Study on the
  effect of fitting distribution}, volume 594.
\newblock Trans Tech Publ, 2014.

\bibitem{noor2015}
Norazian~Mohamed Noor, Mohd~Mustafa Al~Bakri~Abdullah, Ahmad~Shukri Yahaya, and
  Nor~Azam Ramli.
\newblock Comparison of linear interpolation method and mean method to replace
  the missing values in environmental data set.
\newblock In {\em Materials Science Forum}, volume 803, pages 278--281. Trans
  Tech Publ, 2015.

\bibitem{kornelsen2014}
Kurt Kornelsen and Paulin Coulibaly.
\newblock Comparison of interpolation, statistical, and data-driven methods for
  imputation of missing values in a distributed soil moisture dataset.
\newblock {\em Journal of Hydrologic Engineering}, 19(1):26--43, 2014.

\bibitem{hong2020}
Shangzhi Hong and Henry~S Lynn.
\newblock Accuracy of random-forest-based imputation of missing data in the
  presence of non-normality, non-linearity, and interaction.
\newblock {\em BMC medical research methodology}, 20(1):1--12, 2020.

\bibitem{stekhoven2012}
Daniel~J Stekhoven and Peter B{\"u}hlmann.
\newblock Missforest—non-parametric missing value imputation for mixed-type
  data.
\newblock {\em Bioinformatics}, 28(1):112--118, 2012.

\bibitem{wulff2017}
Jesper~N Wulff and Linda Ejlskov.
\newblock Multiple imputation by chained equations in praxis: Guidelines and
  review.
\newblock {\em Electronic Journal of Business Research Methods}, 15(1), 2017.

\bibitem{azur2011}
Melissa~J Azur, Elizabeth~A Stuart, Constantine Frangakis, and Philip~J Leaf.
\newblock Multiple imputation by chained equations: what is it and how does it
  work?
\newblock {\em International journal of methods in psychiatric research},
  20(1):40--49, 2011.

\bibitem{pampaka2016}
Maria Pampaka, Graeme Hutcheson, and Julian Williams.
\newblock Handling missing data: analysis of a challenging data set using
  multiple imputation.
\newblock {\em International Journal of Research \& Method in Education},
  39(1):19--37, 2016.

\bibitem{imputena2020}
Miguel Macarro.
\newblock imputena 1.0.
\newblock \url{https://pypi.org/project/imputena/}, 2020.

\bibitem{missingpy2018}
Ashim Bhattarai.
\newblock missingpy 0.2.0.
\newblock \url{https://pypi.org/project/missingpy/}, 2018.

\bibitem{Saurav2016}
Saurav Kaushik.
\newblock Introduction to feature selection methods with an example (or how to
  select the right variables?
\newblock
  \url{https://www.analyticsvidhya.com/blog/2016/12/introduction-to-feature-selection-methods-with-an-example-or-how-to-select-the-right-variables/},
  2016.

\bibitem{chandrashekar2014}
Girish Chandrashekar and Ferat Sahin.
\newblock A survey on feature selection methods.
\newblock {\em Computers \& Electrical Engineering}, 40(1):16--28, 2014.

\bibitem{subbiah2021}
Siva~Sankari Subbiah and Jayakumar Chinnappan.
\newblock Opportunities and challenges of feature selection methods for high
  dimensional data: A review.
\newblock {\em Ing{\'e}nierie des Syst{\`e}mes d'Information}, 26(1), 2021.

\bibitem{kumar2014}
Vipin Kumar and Sonajharia Minz.
\newblock Feature selection: a literature review.
\newblock {\em SmartCR}, 4(3):211--229, 2014.

\bibitem{saeys2007}
Yvan Saeys, Inaki Inza, and Pedro Larranaga.
\newblock A review of feature selection techniques in bioinformatics.
\newblock {\em bioinformatics}, 23(19):2507--2517, 2007.

\bibitem{schmitt2015}
Peter Schmitt, Jonas Mandel, and Mickael Guedj.
\newblock A comparison of six methods for missing data imputation.
\newblock {\em Journal of Biometrics \& Biostatistics}, 6(1):1, 2015.

\bibitem{li2014}
Yuebiao Li, Zhiheng Li, and Li~Li.
\newblock Missing traffic data: comparison of imputation methods.
\newblock {\em IET Intelligent Transport Systems}, 8(1):51--57, 2014.

\bibitem{botchkarev2018b}
Alexei Botchkarev.
\newblock Performance metrics (error measures) in machine learning regression,
  forecasting and prognostics: Properties and typology.
\newblock {\em arXiv preprint arXiv:1809.03006}, 2018.

\bibitem{christian2018}
Christian Pascual.
\newblock Tutorial: Understanding regression error metrics in python.
\newblock
  \url{https://www.dataquest.io/blog/understanding-regression-error-metrics/},
  2018.

\bibitem{wu2019}
Xuetong Wu, Hadi Akbarzadeh~Khorshidi, Uwe Aickelin, Zobaida Edib, and Michelle
  Peate.
\newblock Imputation techniques on missing values in breast cancer treatment
  and fertility data.
\newblock {\em Health Information Science and Systems}, 7(1):1--8, 2019.

\end{thebibliography}

\end{document}